\documentclass[journal]{IEEEtran}

\usepackage[utf8]{inputenc}
\usepackage{graphicx}
\usepackage{booktabs}
\usepackage{algorithm}
\usepackage{algorithmic}
\usepackage{amsfonts}
\usepackage{multirow}
\usepackage{setspace}
\usepackage{bbding}
\usepackage{amsmath}
\usepackage{xcolor}
\usepackage{color}
\usepackage{colortbl}

\title{Few-Shot Object Detection via Spatial-Channel State Space Model}

	\author{
	Zhimeng Xin,~\IEEEmembership{}
	Tianxu Wu,~\IEEEmembership{}	
	Yixiong Zou,~\IEEEmembership{}	
	Shiming Chen,~\IEEEmembership{}	
	Dingjie Fu,~\IEEEmembership{}	
	and Xinge You,~\IEEEmembership{Senior Member,~IEEE}

	\thanks{		
		Z. Xin is with the School of Cyber Science and Engineering, Huazhong University of Science and Technology, Wuhan 430074, China (e-mail: zhimengxin15@gmail.com).
		
		Y. Zou is with the School of Computer Science \& Technology, Huazhong University of Science and Technology, Wuhan 430074, China (e-mail: yixiongz@hust.edu.cn). 
		
		T. Wu, S. Chen, and D. Fu, and X. You are with the School of Electronic Information and Communications, Huazhong University of Science and Technology, Wuhan 430074, China (e-mail: wutianxu@hust.edu.cn; gchenshiming@gmail.com; dingjiefu@hust.edu,cn; youxg@mail.hust.edu.cn).

	}
}

\begin{document}

\maketitle
\begin{abstract}
	
Due to the limited training samples in few-shot object detection (FSOD), we observe that current methods may struggle to accurately extract effective features from each channel. Specifically, this issue manifests in two aspects: i) channels with high weights may not necessarily be effective, and ii) channels with low weights may still hold significant value. To handle this problem, we consider utilizing the inter-channel correlation to facilitate the novel model's adaptation process to novel conditions, ensuring the model can correctly highlight effective channels and rectify those incorrect ones. Since the channel sequence is also 1-dimensional, its similarity with the temporal sequence inspires us to take Mamba for modeling the correlation in the channel sequence. Based on this concept, we propose a \textbf{S}patial-\textbf{C}hannel \textbf{S}tate Space \textbf{M}odeling (SCSM) module for spatial-channel state  modeling, which highlights the effective patterns and rectifies those ineffective ones in feature channels. In SCSM, we design the Spatial Feature Modeling (SFM) module to balance the learning of spatial relationships and channel relationships, and then introduce the Channel State Modeling (CSM) module based on Mamba to learn correlation in channels. Extensive experiments on the VOC and COCO datasets show that the SCSM module enables the novel detector to improve the quality of focused feature representation in channels and achieve state-of-the-art performance. 
\end{abstract}


\section{Introduction}
Few-shot object detection (FSOD) emerges as a promising solution to detecting objects with limited annotated data~\cite{tmm1,tmm2,tmm3}. This approach closely aligns with the remarkable human ability to recognize new objects from limited examples, making it a precious technique in scenarios where training data is scarce~\cite{smile}. Existing FSOD methods~\cite{defrcn,fsod20242,ecea,fsod20243,aft} utilize pre-trained models from large-scale datasets and fine-tune them with limited labeled data from novel classes, enabling rapid adaptation to the distinct features of these novel classes.

\begin{figure}[t]
	\begin{center}
		%\fontsize{8}{11}\selectfont 
		\includegraphics[width=0.5\textwidth]{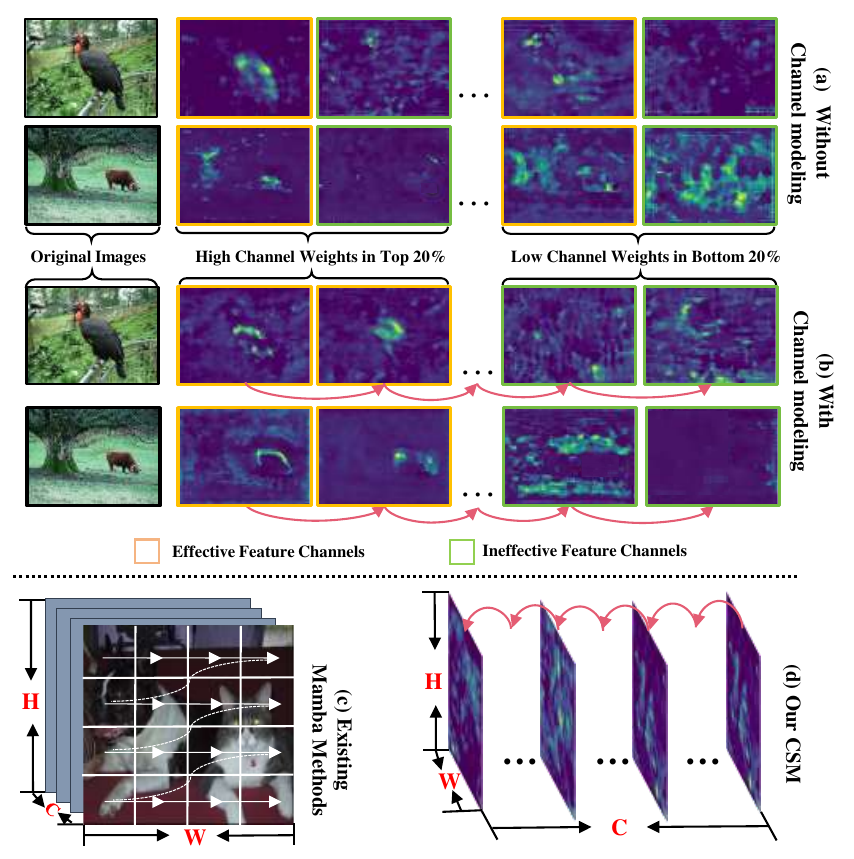}
	\end{center}
	\setlength{\abovecaptionskip}{-0.1cm} 
	\caption{	
		%Channel visualization for the baseline model and ours. Here,  \textcolor{red}{\textbf{C}}: channels, \textcolor{red}{\textbf{W}}: width, and \textcolor{red}{\textbf{H}}: High. (a) and (b) are channel feature representation without and with Mamba state modeling in channels. (c) and (d) are existing mamba-based methods and ours. 			
		Channel visualization for the baseline model and ours. We find the high-weight channels are not really effective as green-boxed channels cannot reflect input objects (a), and vice versa, which leads to the ineffective features extracted from novel-class samples. To handle this problem, we model the correlation between channels to highlight effective patterns and rectify the ineffective ones. 
		We view channels as a 1-dimensional sequence (d), and take Mamba for the channel sequence modeling, different from current works only modeling the spatial patch sequence (c).
		%\textcolor{red}{\textbf{C}}: channels, \textcolor{red}{\textbf{W}}: width, and \textcolor{red}{\textbf{H}}: Height.
		By applying our method, the model can correctly highlight effective channels (b), improving FSOD performance.		
	}

	\label{motivation}
	%\vspace{-0.3cm} 
\end{figure}

%General object detection, trained with abundant samples, can thoroughly learn a diverse range of class features and their intrinsic relationships, which allows it to create a more accurate and effective representation of channel features.
%However, since base and novel classes are non-overlapping, there is a significant semantic gap between base and novel classes~\cite{vfa,metarcnn,fsce}. This gap makes the model tend to extract ineffective or redundant channel features when dealing with novel classes.

%AHowever, since the base and novel classes are non-overlapping, there are significant differences in their data distributions. By this, the prior knowledge from the base model may not assist the novel model in extracting effective and non-redundant channel features in data-scarce scenarios.  

However, due to the limited training samples in novel classes, we find that current works \cite{fsod20242,ecea,fsod20243,aft,snida} may not correctly extract effective features in each channel\footnote{Notably, \textit{traditional object detection}, trained with extensive data, thoroughly learns diverse class features and their relationships, enabling the model to create accurate and effective channel feature representations.}. To illustrate this issue, we visualize a set of feature channels in Fig.~\ref{motivation}. Here, we employ a model trained on base classes and then finetuned on novel classes to extract feature for novel-class samples, and obtain the channel weights using SENet~\cite{senet} in testing. From Fig.~\ref{motivation}(a), we can see the high-weight channels extracted by the existing method~\cite{ecea} without channel modeling may not be really effective. For example, the green-boxed channels majorly contain noisy patterns that can not reflect the object in the image. In contrast, the low-weight channels may not be really ineffective, since orange-boxed channels still capture the object information. As a result, the extracted features on novel classes tend to be ineffective, since the effective channels cannot be correctly highlighted in the extracted features. This indicates that it is hard for the model to fully transfer patterns learned on base classes and adapt them to represent novel classes, leading to low performance in FSOD.

To handle this problem, we consider utilizing the inter-channel correlation to facilitate the base model's adaptation process to novel conditions, ensuring the model can correctly highlight effective channels and rectify those incorrect ones.
To achieve this goal, we view channels as a 1-dimensional sequence. Since the location of each channel is fixed for all inputs given a trained model, the change of channel along the sequence dimension could show similar patterns across samples. 
\textbf{This characteristic of the channel sequence is similar to the image sequence with temporal patterns, which inspires us to take Mamba~\cite{mamba}, as a variant of temporal networks, to model the channel sequence and capture the correlation between channels.}
Unlike current works that usually model the spatial patch sequence, for our task, we view each channel in the sequence as a state, and Mamba is further applied in the modeling of channel sequences, as shown in Fig.~\ref{motivation}(d).

Building upon this concept, we propose the \textbf{S}patial-\textbf{C}hannel \textbf{S}tate Space \textbf{M}odeling (SCSM) module, a variant of Mamba for spatial-channel-sequence modeling, 
%that highlights the correctly transferred patterns and rectifies those incorrect ones in channels to alleviate the semantic gap between the base and novel classes. 
to assist the learning of channels by learning the channel correlation, thereby improving the knowledge transfer and few-shot adaptation.
Specifically, inspired by the spatial-channel attention mechanism \cite{cbam,eca}, we first introduce Spatial Feature Modeling (SFM) in SCSM, which utilizes the multi-head attention mechanism~\cite{self-attention} 
%to ensure that the subsequent extracted channel features are valid and balance the learning of state relationships among subsequent channel features~\cite{cbam}. 
for modeling the spatial correlation and balancing the learning of spatial relationships and channel relationships~\cite{cbam}.
We then design Channel State Modeling (CSM) based on Mamba to learn the correlation in channels. By modeling the channel sequence, our model can correctly highlight effective channels, as shown in Fig.~\ref{motivation}(b) where the high-weight channels are all representative of the input object. 
Extensive experiments on the PASCAL VOC and COCO datasets show that the SCSM module enables the novel detector to improve the quality of focused feature representation in channels and improve the performance of FSOD.

Our contributions can be summarized as follows:

\begin{itemize} 
	 
	\item We propose the SCSM module as a variant of Mamba for spatial-channel-sequence modeling. To the best of our knowledge, we are the first to take Mamba to model the correlation in channel sequences for the FSOD task.
		
	\item In the SCSM module, we introduce CSM to capture the correlations among channels for highlighting effective channels. Furthermore, we design SFM to balance the learning of spatial correlation and channel correlation.
	
	\item Extensive experiments on the VOC and COCO datasets show that the SCSM module enables the novel detector to improve the quality of focused feature representation in channels and improve the performance of FSOD.
	
\end{itemize}

\section{Related Work}

\subsection{Few-Shot Object Detection}
Currently, following the principles of transfer learning, most FSOD methods often adopt knowledge transferred from classes with abundant base data to train the novel model using only a few annotated samples \cite{xinservey,DC,EME,PCD}. Based on this fact, two main approaches are commonly used: meta-learning-based methods \cite{fsod20241,fsod20243} and two-stage fine-tuning-based methods \cite{fsod20242}. Meta-learning-based FSOD methods \cite{metafrcn,metarcnn,qa} divide the dataset into a series of episode tasks and learn general knowledge or patterns from these tasks to generalize to new tasks. However, such methods typically involve complex training processes and architectural designs \cite{TFA}. To address this issue, two-stage fine-tuning methods require training only one task in the base and novel phases, then fine-tuning the novel detector, achieving performance that matches or surpasses complex meta-learning methods \cite{defrcn,SRR,niff,snida}. For example, DeFRCN \cite{defrcn} achieves rapid learning of novel models by simply fine-tuning the gradient backward between the backbone and detection head, enabling independent optimization at different modules. NIFF \cite{niff} is proposed to alleviate forgetting without base data in the novel stage. By decoupling foreground and background, SNIDA \cite{snida} increases their diversity, further improving the performance of FSOD. Nonetheless, such two-stage fine-tuning-based methods lack a uniform fine-tuning strategy.

\subsection{Feature Correlation for FSOD}
Due to significant differences in feature distributions between novel and base classes, the aforementioned FSOD methods face domain shift issues to varying degrees. In scenarios with extremely limited samples, models struggle to learn intra-class and inter-class feature representations effectively, resulting in classification confusion. To address this challenge, contrastive \cite{fsce} and metric learning \cite{zhang,repmet,metricfsod2} techniques alleviate feature cognition ambiguity by measuring intra-class and inter-class feature similarity. Furthermore, VAE-based approaches \cite{vfa,vae2} aim to aggregate effective features of classes to enhance feature representation. However, when dealing with very few samples (1 or 2 shots) and a large number of classes, such as in the case of COCO with 20 novel classes, models tend to overly emphasize the differences between classes, neglecting subtle variations within categories. This intra-class bias increases uncertainty in the model's recognition of intra-class features and may worsen classification confusion. To tackle this issue, transformer-based methods \cite{fct,metadetr,ecea} enhance awareness of intra-class information by focusing on correlations between local features. Nevertheless, these methods do not model overall shapes or directly infer global shapes. For instance, ECEA \cite{ecea} can infer correlations between unseen local features and known local features through extensible learning, but strong correlations between different instances of the same category may be treated as a single one.

\begin{figure*}[t]
	\begin{center}
		%\fontsize{8}{11}\selectfont 
		\includegraphics[width=0.95\textwidth]{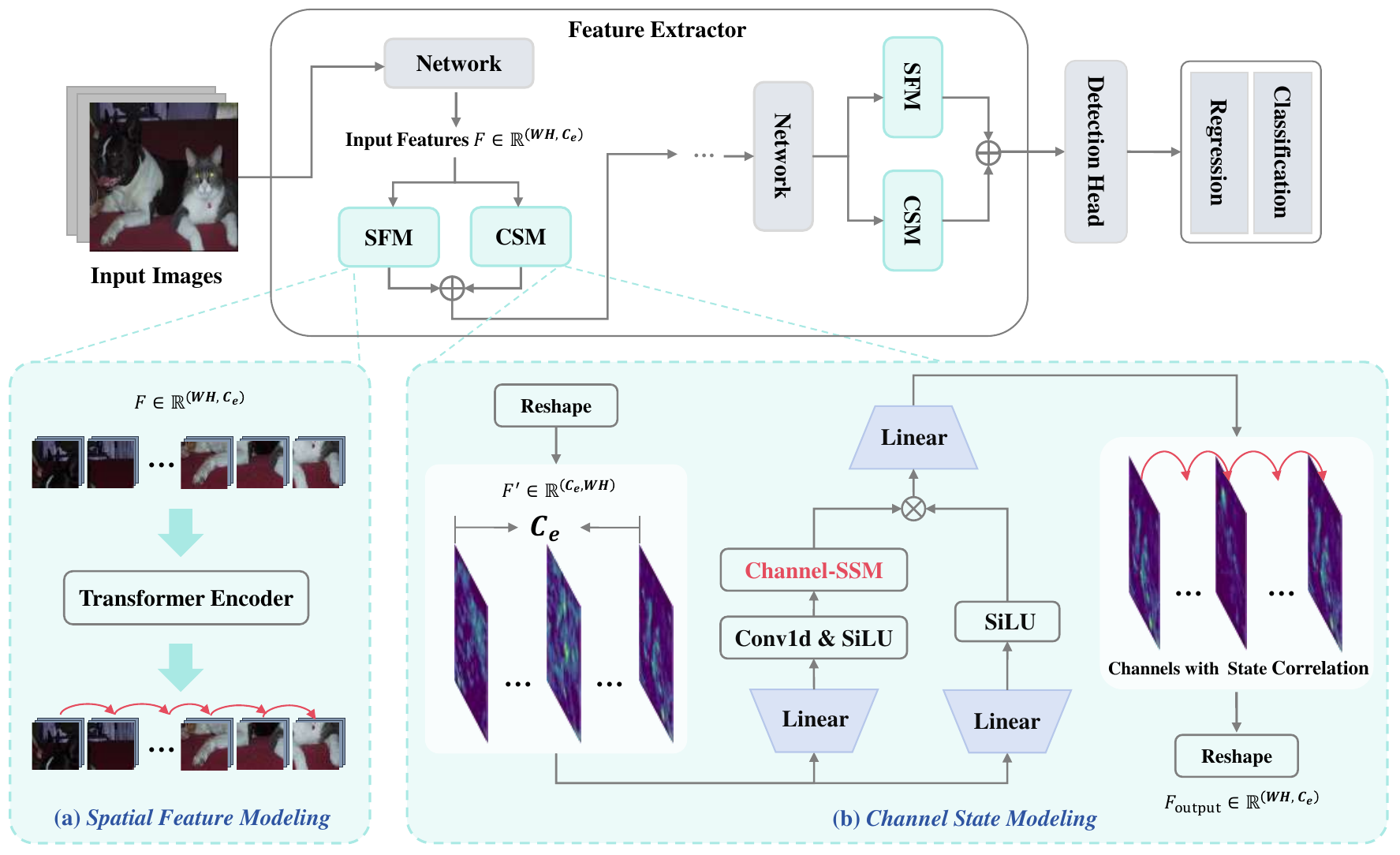}
	\end{center}
	\setlength{\abovecaptionskip}{-0.1cm} 
	\caption{The framework of our proposed SCSM. Our SCSM module includes SFM and CSM components that are parallelly inserted after each backbone stage. (a) SFM employs the multi-head attention mechanism to learn the spatial feature correlation, ensuring balanced learning of state relationships among subsequent channel features. (b) In CSM, we construct a batch feature matrix consisting of $\boldsymbol{C_e}$ sequences, where each sequence is represented by a $\boldsymbol {P}$-dimensional feature vector. Here, we assume that the batch of the input features $F$ is 1. 
	}
	\vspace{-0.3cm} 
	\label{framework}
	%\vspace{-0.2cm} 
\end{figure*}

\subsection{State Space Model}

Compared to transformer-based state sequence models, the State Space Model (SSM) based mamba \cite{mamba,mamba2} introduces selective mechanisms and temporal variability to adaptively adjust state space parameters, optimizing computational efficiency and memory usage. Mamba, with its efficient sequential modeling capability, has been successfully applied in the field of computer vision \cite{videomamba, mambasurvey}. For instance, Zhu \textit{et al.} \cite{vim} proposed Vision Mamba (ViM), a visual representation learning model similar to ViTs. Vim combines bidirectional state space models with positional embeddings to handle various visual tasks. Liu et al. \cite{vmamba} introduced VMamba, which acts as a backbone to improve computational efficiency and visual representation learning performance by combining state space models and selective scanning mechanisms. LocalMamba \cite{localmamba} improves the efficiency and performance of image representation through windowed selective scanning. However, such methods model sequences based on spatial features, neglecting the quality of feature expression within channels.

\subsection{Channel Attention} Channel attention dynamically adjusts the importance of different channels by exploiting relationships between features, thereby enhancing the performance of image processing tasks. For instance, SENet \cite{senet} compresses channel information into a 1-dimensional vector through global average pooling, then obtains the weight of each channel through fully connected layers and a Sigmoid activation function to enhance crucial features. CBAM \cite{cbam} combines channel attention and spatial attention mechanisms to address the lack of consideration for spatial features in SENet, yet the feature enhancement in two dimensions introduces additional overhead. ECA \cite{eca} reduces computational complexity by eliminating fully connected layers and directly performing 1-dimensional convolution operations in the channel dimension, while maintaining the effectiveness of the channel attention mechanism. In contrast to methods that use a scalar to represent channels, FcaNet \cite{fcanet} effectively enhances model performance through multi-spectral channel attention, further strengthening the channel attention mechanism. However, these approaches overlook global modeling. GCNet \cite{gcnet} utilizes global contextual information to generate channel attention weights, thereby boosting model performance and implementing global context modeling to enhance channel attention. Unfortunately, global context information is obtained through global average pooling, a process conducted in the spatial dimension, but the final attention weights are applied in the channel dimension. Our SCSM module achieves long-range modeling between channels, further enhancing channel feature representation.

\section{Methodology}

In this section, we begin by presenting the preliminary definitions of FSOD and Mamba. We then introduce the innovative SCSM module to highlight the correctly transferred patterns and rectify those incorrect ones in channels.
% In SCSM, we design SFM to balance the learning of state relationships among subsequent channel features and CSM to capture the relationships between feature states in channels.


\subsection{Preliminary Definition}

\textbf{Task Definition.}
According to the existing definition of FSOD, denote  $\mathbb{D} = \{(x, y) \mid x \in \mathcal{X}, y \in \mathcal{Y}\}$ as a large-scale dataset, where $x$ represents the input image, and $y = \{l_i, b_i\}_{i=1}^K$ represents the corresponding manual annotation information, including the class label $l$ and its bounding box $b$. We divide $\mathbb{D}$ into a fully annotated base dataset $\mathbb{D}_b$ with class set $\mathbb{C}_b$ and a sparsely annotated novel dataset $\mathbb{D}_n$ with class set $\mathbb{C}_n$, typically containing few samples, where $\mathbb{C} = \mathbb{C}_b \cup \mathbb{C}_n$ and $\mathbb{C}_b \cap \mathbb{C}_n = \emptyset$. We adopt a two-stage fine-tuning paradigm for training. In the first stage, we train an initial model $\mathcal{M}_{\text{init}}$ using $\mathbb{D}_b$ to obtain a base model $\mathcal{M}_{\text{base}}$. In the novel stage, we train a novel model $\mathcal{M}_{\text{fsod}}$ using $\mathbb{D}_n$. Furthermore, if the dataset $\mathbb{C}_b \cup \mathbb{C}_n$ forms a balanced dataset $\mathbb{D}_f$ containing only a few annotated classes $\mathbb{C}$ in the novel stage, it is referred to as generalized few-shot object detection (G-FSOD).

\textbf{Preliminary Mamba.}
Recently, the SSM-based Mamba, based on structured state space sequence models (S4), takes inspiration from a continuous system \cite{mamba,mamba2}. In this system, the hidden state $h(t)$ lies in the real space of dimension $\boldsymbol{C}$, while the function or sequence $x(t)$ maps from the real numbers to $y(t)$ in the real space. The system employs $\mathcal{A}$ (evolution parameter) and $\mathcal{B}$ (projection parameter), both of which are matrices, $\mathcal{A} \in \mathbb{R}^{\boldsymbol{C} \times \boldsymbol{C}}$ and $\mathcal{B} \in \mathbb{R}^{\boldsymbol{C} \times 1}$. Mamba serves as the discrete counterpart to the continuous system, using the zero-order hold (ZOH) technique to convert continuous parameters $\mathcal{A}$ and $\mathcal{B}$ to discrete parameters $\mathcal{A}_d$ and $\mathcal{B}_d$, whcih can be given by
\begin{equation}
	\begin{aligned} 
		\begin{array}{l}
			\Delta t = \frac{1}{\sqrt{\lambda_{\text{max}}}}\\
			\mathcal{A}_d = \exp(\Delta t \cdot \mathcal{A}) \\
			\mathcal{B}_d = (\Delta t \cdot \mathcal{A})^{-1}(\exp(\Delta t \cdot \mathcal{A}) - I) \cdot \mathcal{B},
		\end{array}
	\end{aligned} 
	\label{AB}
\end{equation}
where $\Delta t$ is a time scale parameter, typically calculated based on the maximum eigenvalue $\lambda_{\text{max}}$ of the state matrix $\mathcal{A}$. Thus, the discretized Mamba model can be given by
\begin{equation}
	\begin{array}{l}
		%h_t=\bar{A} h_{t-1}+\bar{B} x_t \\
		%y_t=C h_t
		h_t=\mathcal{A}_d h_{t-1}+\mathcal{B}_d x_t \\
		y_t=\mathcal{C} h_t.
	\end{array}	
\end{equation}

\subsection{Spatial-Channel State Space Modeling Module}

%Due to the different data sources, there exists a significant semantic gap in features between novel and base classes \cite{vfa,metarcnn,fsce}. This gap leads to the novel model either extracting insufficient or redundant channel features. 
Due to the limited availability of data samples, the novel model tends to extract ineffective or redundant channel features when dealing with novel classes.
To mitigate this issue, we propose an SCSM module that models the long-term dependencies between channel states within high-quality spatial features. This module enables the novel model to accurately highlight effective channels while correcting inaccurate ones.

%can assist the novel model in leveraging class-specific features effectively while filtering out irrelevant noise.

%To address these challenges, we propose the SCSM module as a plug-and-play component that serves as a novel feature extractor. It is strategically positioned after each backbone stage.

Specifically, as illustrated in Fig. \ref{framework}, the SCSM module is designed as a residual block \cite{resnet}, which is inserted after each stage of the backbone, thereby enhancing the model’s capability to tackle few-shot tasks effectively. This integration allows Mamba, capable of capturing global dependencies across long sequential channels, to bolster feature extractors that are limited to local feature representation, such as ResNet. Consequently, this enhances the model's overall feature representation. Furthermore, in the base training phase, both the backbone network and the SCSM module are trained concurrently without freezing any parameters. During the novel phase, we freeze the backbone while allowing the SCSM module to remain trainable, refraining from fine-tuning any other parameters. This strategy significantly reduces the time required for fine-tuning the novel model.

%Finally, as illustrated in Figure \ref{framework}, we position our SCSM module after each feature extraction block, thereby enhancing the model’s capability to tackle few-shot tasks effectively.

\subsection{Spatial Feature Modeling}
To enhance the model's ability to capture the correlation between feature channels, we design SFM, as shown in Fig. \ref{framework}(a). Specifically, taking stage 5 of ResNet101 as an example, we extract the features from the final convolutional layer, resulting in a batch image feature tensor with shape $(\boldsymbol{B}, \boldsymbol{C}, \boldsymbol{W}, \boldsymbol{H})$. We then permute and reshape this tensor to $(\boldsymbol{S},\boldsymbol{B}, \boldsymbol{C_e})$, where $\boldsymbol{S} = \boldsymbol{W} \times \boldsymbol{H}$. Subsequently, we compress the channels through Convd2 from $\boldsymbol{C}$ to $\boldsymbol{C_e}$ to alleviate the computational complexity. The condensed feature maps $(\boldsymbol{S}, \boldsymbol{B}, \boldsymbol{C_e})$ are utilized for space modeling. Denote $f \in \mathbb{R}^{ \boldsymbol{S} \times \boldsymbol{B} \times \boldsymbol{C_e}}$ as the query sequence of spatial feature patches. The spatial features can be modeled as

\begin{equation}
	\begin{aligned} 
		F(f)=\sum_{n=1}^{\boldsymbol{S}} \frac{e^{f \cdot \mathcal{W}^{q }\left(f_q \mathcal{W}_n^k\right)^T}}{\sum_{n=1}^{\boldsymbol{S}} e^{f \cdot \mathcal{W}^q \left(f_q \mathcal{W}_n^k\right)^T}}  f \cdot \mathcal{W}_n^v,
	\end{aligned}
\end{equation}  
where $\mathcal{W}^q$, $\mathcal{W}_n^v$, and $\mathcal{W}_n^k$ represent three different weight matrices, $n$ is n-th spatial feature patch. Following the transformer-based work \cite{swin,self-attention}, we conduct multiple-head feature attention to enhance further the relationship between spatial feature patches $f$, which can be given by

\begin{equation}
	F=\sum_{m=1}^M  F (f) \mathcal{W}_m,
\end{equation}
where $\mathcal{W}_m$ represents weight vectors aggregation and $F$ is modeled spatial features. 

% $\mathcal{X}\mathcal{Y}$ $\boldsymbol{C}$

\begin{algorithm}[t]
	\caption{Channel State Modeling Algorithm}
	\label{A1}
	%\vspace{-0.6cm}
	\begin{algorithmic}[1]
		\REQUIRE Input channel features  $F \in \mathbb{R}^{\boldsymbol {B} \times \boldsymbol {S} \times \boldsymbol {C_e}}$
		\ENSURE Output channel features  $F_{\text{output}} \in \mathbb{R}^{\boldsymbol {B} \times \boldsymbol {S} \times \boldsymbol {C_e}}$

		\STATE \textbf{Feature Channel Initializion}:

		\STATE $F' \gets \text{Permute}(F)$ \COMMENT{$F' \in \mathbb{R}^{\boldsymbol {B} \times \boldsymbol {C_e} \times \boldsymbol {S}}$}
		
		\STATE $T \gets \text{DownSampling}(F')$ \COMMENT{$T \in \mathbb{R}^{\boldsymbol {B} \times \boldsymbol {C_e} \times \boldsymbol {P}}$}
		
		\STATE \textbf{Input Processing}: %Normalize the input channel 
		\STATE $T \gets \text{Norm}(T)$ \COMMENT{$T \in \mathbb{R}^{\boldsymbol {B} \times \boldsymbol {C_e} \times \boldsymbol {P}}$}
		
		%\STATE \textbf{Linear Transformation}: Generate intermediate feature channel representations
		\STATE $X \gets \text{Linear}_1(T)$ \COMMENT{$X \in \mathbb{R}^{\boldsymbol {B} \times \boldsymbol {C_e} \times \boldsymbol {D}}$}
		\STATE $Z \gets \text{Linear}_2(T)$ \COMMENT{$Z \in \mathbb{R}^{\boldsymbol {B} \times \boldsymbol {C_e} \times \boldsymbol {D}}$}
		
		%\STATE \textbf{Processing}: Process feature channel 
		
		\STATE $\mathcal{B} \gets \text{Linear}'(X)$ \COMMENT{$B \in \mathbb{R}^{\boldsymbol {B} \times \boldsymbol {C_e} \times \boldsymbol {D}}$}		
		
		\STATE $\mathcal{C} \gets \text{Linear}(X)$ \COMMENT{$\mathcal{C} \in \mathbb{R}^{\boldsymbol {B} \times \boldsymbol {C_e} \times \boldsymbol {D}}$}

		\STATE \textbf{Channel State Space Model}:

		\STATE $\mathcal{A} \gets \text{State Matrix}$ \COMMENT{$\mathcal{A} \in \mathbb{R}^{\boldsymbol {D} \times \boldsymbol {D}}$}
		%\STATE $\Delta t \gets \frac{1}{\sqrt{\lambda_{\text{max}}}}$ %\COMMENT{Time step based on the maximum eigenvalue of $A$}
		%\STATE $\mathcal{A}_d \gets \exp(\Delta t \cdot \mathcal{A})$
		%	\STATE $\mathcal{B}_d \gets (\Delta t \cdot \mathcal{A})^{-1}(\exp(\Delta t \cdot \mathcal{A}) - I) \cdot \mathcal{B}$
		
		\STATE $\mathcal{A}_d, \mathcal{B}_d \gets Eq. (\ref{AB})$

		\STATE $X \gets \text{SiLU}(\text{Conv1d}(X))$ \COMMENT{$X \in \mathbb{R}^{\boldsymbol {B} \times \boldsymbol {C_e} \times \boldsymbol {D}}$}
		%\STATE $A_d \gets e^{A \Delta t}$
		%\STATE $B_d \gets A^{-1}(e^{A \Delta t} - I)B$
		\STATE $y \gets \text{SSM}(A_d, B_d, C)(X^2)$ \COMMENT{$y \in \mathbb{R}^{\boldsymbol {B} \times \boldsymbol {C_e} \times \boldsymbol {D}}$}
		
		%\STATE \textbf{Get Gated $y$}:
		\STATE $y \gets y \cdot \text{SiLU}(Z)$ \COMMENT{$y \in \mathbb{R}^{\boldsymbol {B} \times \boldsymbol {C_e} \times \boldsymbol {D}}$}
		\STATE $y \gets\text{Linear}_3(y)$ \COMMENT{$y \in \mathbb{R}^{\boldsymbol {B} \times \boldsymbol {C_e} \times \boldsymbol {P}}$}	
		
		\STATE \textbf{Feature Restoration}:
		
		\STATE $F' \gets \text{UpSampling}(y)$ \COMMENT{$F' \in \mathbb{R}^{\boldsymbol {B} \times \boldsymbol {C_e} \times \boldsymbol {S} }$}

		\STATE $F' \gets \text{Permute}(F')$ \COMMENT{$F' \in \mathbb{R}^{\boldsymbol {B} \times \boldsymbol {S} \times \boldsymbol {C_e}}$}
		
		%	\STATE \textbf{Residual Connection}
		\RETURN $F_{\text{output}} \gets F' + F$ %\COMMENT{$_{\text{output}} \in \mathbb{R}^{\boldsymbol {B} \times \boldsymbol {C_e} \times \boldsymbol {P}}$}		
		%\RETURN $F_{\text{output}}$
	\end{algorithmic}	
	
\end{algorithm}

\subsection{Channel Sate Modeling}

Inspired by the memory-capacity learning of Mamba \cite{mamba,vim}, we introduce a novel correlation learning strategy based on channel features. This strategy effectively captures class-specific global features within long sequences while filtering out irrelevant noise. The process of our proposed CSM approach is outlined in Algorithm \ref{A1} and Fig. \ref{framework}(b).

Specifically, Mamba or Mamba-based vision works are designed for spatial feature sequences, with the Mamba block being created based on the channel size. Unlike such models, we treat channels as sequences in CSM. To achieve this, we first downsample the feature map $F \in \mathbb{R}^{\boldsymbol{S} \times \boldsymbol{B} \times \boldsymbol{C_e}}$ from SFM and permute it, resulting in $F \in \mathbb{R}^{\boldsymbol {B} \times \boldsymbol {C_e} \times \boldsymbol {P}}$, where $\boldsymbol{P}$ represents the size of the spatial features after downsampling.

\begin{figure}[t]
	\begin{center}
		%\fontsize{8}{11}\selectfont 
		\includegraphics[width=0.5\textwidth]{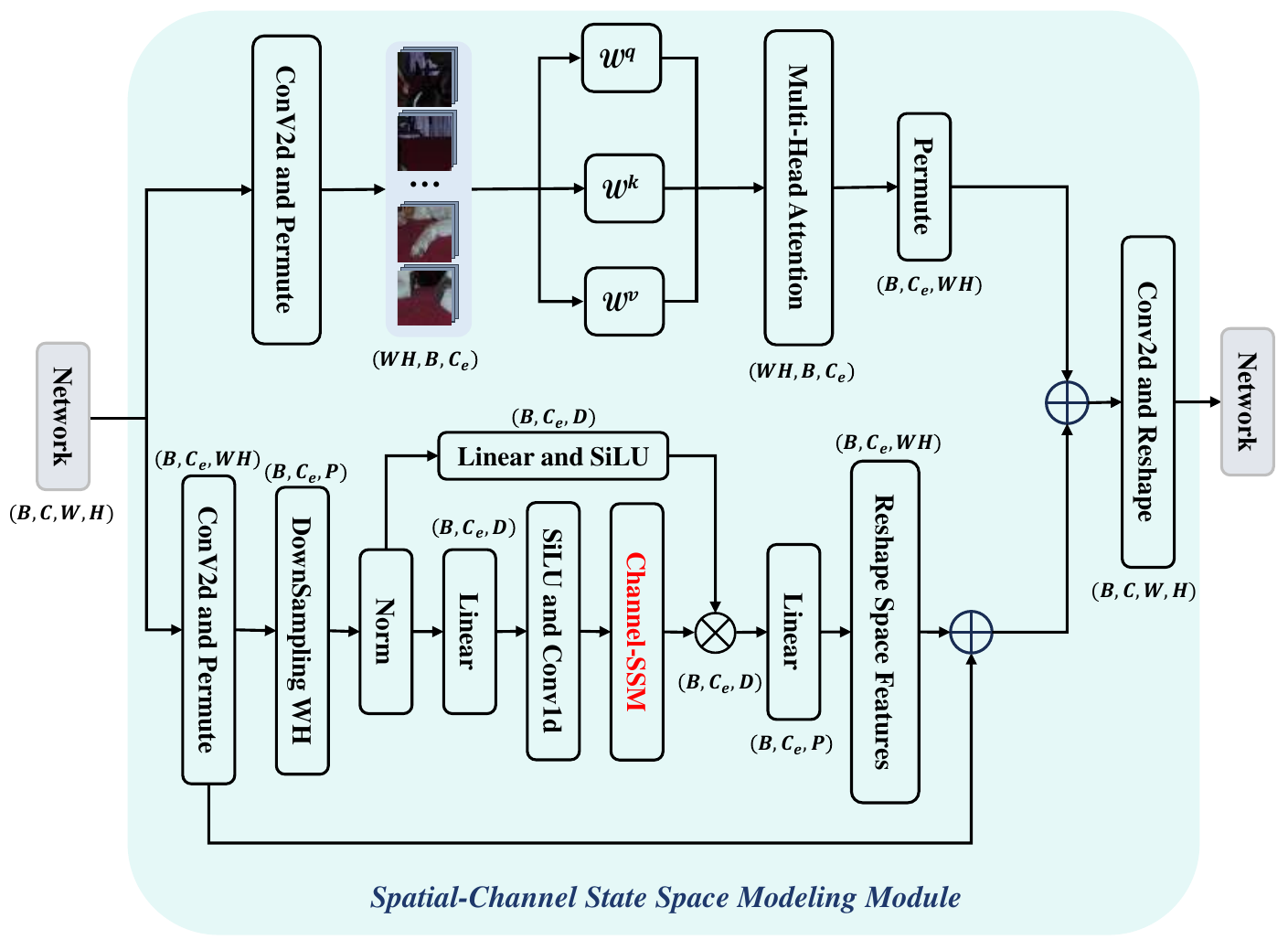}
	\end{center}
	\setlength{\abovecaptionskip}{-0.1cm} 
	\caption{Variations of input feature size in SCSM.
	}
	\vspace{-0.3cm} 
	\label{framework-b}
	%\vspace{-0.2cm} 
\end{figure}

Building upon this, we construct a Mamba block with a  $\boldsymbol{P}$-dimensional feature vector. Here, the channels $\boldsymbol{C_e}$ in feature $F$ are treated as a sequence, allowing the model to capture the variations in feature states and global features pertinent to the learning task. To facilitate subsequent work, the channel sequence $\{ c_1, c_2...c_n \}$ $ \in \boldsymbol{C_e}$ is normalized and linearly projected onto a vector of size $\boldsymbol{D}$ by $\mathcal{W_x}$ and $\mathcal{W_z}$ weights to obtain

\begin{equation}
	\begin{aligned}
		X & =\left[c_1 \mathcal{W_x} ; c_2 \mathcal{W_x} ; \cdots ; c_n\mathcal{W_x}\right] \\
		Z & =\left[c_1 \mathcal{W_z} ; c_2 \mathcal{W_z} ; \cdots ; c_n \mathcal{W_z}\right].
	\end{aligned}
\end{equation}
Then, we linearly transform the acquired $X$ into continuous matrices $\mathcal{B}$ and $\mathcal{C}$. We initialize matrix $\mathcal{A}$ using Highly Optimized Polynomial Projection Operator (HiPPO) matrix \cite{hippo}. Subsequently, we discretize matrices $\mathcal{A}$ and $\mathcal{B}$ to obtain $\mathcal{A}_d$ and $\mathcal{B}_d$ for the following channel-SSM learning phase of the model. After the application of Conv1d on $X$, we introduce non-linear traits using SiLU, which enhances the expressiveness of the model. Following a single iteration of the channel-SSM, we obtain the output $y$. The output $y$ is then gated by SiLU($Z$) and linearly mapped to restore it to $\boldsymbol{P}$ dimensions. Finally, the feature dimension $\boldsymbol{P}$ of $y$ is upsampled and permuted to match the feature size $F'$ of SFM, as shown in in Algorithm \ref{A1}. $F'$ is then concatenated with the residual feature $F_{\text{output}}$. The details on the variations in SCSM feature sizes can be seen in Fig. \ref{framework-b}.

By leveraging this approach, the few-shot model can effectively learn category-specific global features within the feature channels while filtering out noise that is not relevant to the classes in the data with very scarcity scenarios. This aids the model to improve the knowledge transfer and few-shot adaptation.  Moreover, the model is capable of adapting to changes in feature states based on different contexts, allowing it to effectively identify objects even when they exhibit different states in alternative scenarios.

\begin{table*}[t]
	\centering	
	\tabcolsep=0.09cm
	
	\caption{Performance comparison among SCSM and mainstream FSOD methods based on PASCAL VOC with three random novel splits. Bold font indicates the SOTA result in the group. 
	Symbol `*' represents the results are reported by ours and  Swin-B is the backbone of Swin Transfromer with base size.
	% Symbol `-' represents unreported results in the original work.
	}
	\vspace{-0.3cm}
	\scalebox{1}{
		\begin{tabular}{@{}l|c|ccccc|ccccc|ccccc|c@{}}
			\toprule
			 \multirow{2}{*}{Methods/shots} & \multirow{2}{*}{Backbone} & \multicolumn{5}{c|}{Novel Split1}                                                                                    & \multicolumn{5}{c|}{Novel Split2}                                                                                    & \multicolumn{5}{c|}{Novel Split3}   & \multirow{2}{*}{Avg.}                                                                                                      \\ 
			&	&                   \multicolumn{1}{c}{1}                         & \multicolumn{1}{c}{2}                         & \multicolumn{1}{c}{3}                         & \multicolumn{1}{c}{5}    & \multicolumn{1}{c|}{10}    & \multicolumn{1}{c}{1}                         & \multicolumn{1}{c}{2}                         & \multicolumn{1}{c}{3}                         & \multicolumn{1}{c}{5}    & \multicolumn{1}{c|}{10}     & \multicolumn{1}{c}{1}                         & \multicolumn{1}{c}{2}                         & \multicolumn{1}{c}{3}                         & \multicolumn{1}{c}{5}    & \multicolumn{1}{c|}{10}                        \\ \midrule

		%	TFA w/ fc~\cite{TFA} & ResNet101 & 36.8 & 29.1 & 43.6 & 55.7 &57.0 &18.2 &29.0 &33.4 &35.5 &39.0 &27.7 &33.6 &42.5 &48.7 &50.2 & 38.7 \\
		
	%	 FSRW~\cite{fsrw} & YOLOv2 & 14.8  & 15.5  & 26.7  & 33.9 &  47.2 &  15.7  & 15.3  & 22.7  & 30.1 &  40.5  & 21.3  & 25.6  & 28.4  & 42.8 &  45.9  & 28.4 \\
		
	MetaDet~\cite{metadet} & VGG-16                        &  18.9                      & 20.6                      & 30.2                      & 36.8 & 49.6 & 21.8                      & 23.1                      & 27.8                      & 31.7  & 43.0 & 20.6                      & 23.9                      & 29.4                      & 43.9                      & 44.1           &    31.0         \\ 
		
				TFA w/ cos~\cite{TFA}    & ResNet101                       &  39.8                      & 36.1                      & 44.7                      & 55.7 & 56.0 & 23.5                      & 26.9                      & 34.1                      & 35.1 & 39.1 & 30.8                      & 34.8                      & 42.8                      & 49.5                      & 49.8       & 39.9               \\

	%			Meta R-CNN~\cite{metarcnn}   & ResNet101                  &  19.9                      & 25.5                      & 35.0                      & 45.7 & 51.5 & 10.4                      & 19.4                      & 29.6                      & 34.8 & 45.4 & 14.3                      & 18.2                      & 27.5                      & 41.2                      & 48.1         & 31.1             \\ 

	%		FSCE~\cite{fsce}                       & ResNet101 &  44.2                      & 43.8                      & 51.4                     & 61.9 & 63.4 & 27.3                      & 29.5                      & 43.5                      & 44.2 & 50.2 & 37.2                      & 41.9                      & 47.5                      & 54.6                      & 58.5            &    46.6       \\ 
			
			%	QA-FewDet~\cite{qa} & ResNet101 &  42.4 &51.9 &55.7 &62.6 &63.4 &25.9 &37.8 &46.6 &48.9 &51.1 &35.2 &42.9 &47.8 &54.8 &53.5 & 48.0\\
			
	%			Meta FRCN~\cite{metafrcn}& ResNet101 & 43.0 & 54.5 &60.6 &66.1 &65.4 &27.7 &35.5 &46.1 &47.8 &51.4 &40.6 &46.4 &53.4 & 59.9 &58.6  & 50.5 \\

			%CME \cite{cme}& ResNet101  &41.5 &47.5 &50.4 &58.2 &60.9 &27.2 &30.2 &41.4 &42.5 &46.8 &34.3 &39.6 &45.1 &48.3 &51.5 & 44.4 \\
			
			% Halluc \cite{hall}& ResNet101   & 47.0  & 44.9  & 46.5  & 54.7  & 54.7  & 26.3  & 31.8  & 37.4  & 37.4  & 41.2  & 40.4  & 42.1  & 43.3  & 51.4  & 49.6 & 43.2 \\
			
				FCT~\cite{fct}& PVTv2 &49.9 &57.1 &57.9 &63.2 & 67.1 &27.6 &34.5 &43.7 &49.2 &51.2 &39.5 &54.7 &52.3 &57.0 &58.7 & 50.9\\
			
			Meta-DETR~\cite{metadetr}& ResNet101 &35.1 &49.0 &53.2 &57.4 &62.0 &27.9 &32.3 &38.4 &43.2 &51.8 &34.9 &41.8 &47.1 &54.1 &58.2 & 45.8 \\

			 % CoCo-RCNN  \cite{eccv2022}& ResNet101    &  33.5   &  44.2   &  50.2   &  57.5   &  63.3   &  25.3   &  31.0   &  39.6   &  43.8   &  50.1   &  24.8  &   36.9  &   42.8  &   50.8  &   57.7  & 43.4\\
			
			 VFA \cite{vfa}& ResNet101 & 47.4 & 54.4 & 58.5 & 64.5 & 66.5 & 33.7 & 38.2 & 43.5 & 48.3 & 52.4 & 43.8 & 48.9 & 53.3 &  58.1 & 60.0 &  51.4	 \\

			FPD \cite{fpd}& ResNet101 &   48.1   &  62.2   &  64.0   &  67.6   &  68.4   &  29.8   &  43.2   &  47.7   &  52.0   &  53.9   &  44.9   &  53.8   &  58.1   &  61.6   &  62.9 & 54.6 \\
			
		%	  DC \cite{DC}& ResNet101 & 56.6 &  59.6  & 62.9 &  65.6  & 62.5  & 29.7  & 38.7  & 46.2  & 48.9  & 48.1 &  47.9 &  51.9 &  53.3  & 56.1  & 59.4 & 52.5\\
			
		%	PDC \cite{PCD}& ResNet101 & 41.6  & 44.2  & 50.8  & 56.7  & 61.5  & 31.1  & 30.6  & 41.1  & 43.1  & 48.6  & 34.6  & 43.0  & 43.0  & 49.5  & 52.2 & 44.8 \\

		% EME \cite{EME} & ResNet101& 43.9  & 47.9  & 53.7  & 62.1  & 62.9  & 26.0  & 30.1  & 45.7  & 46.3  & 52.0  & 38.2  & 49.6 &  51.1  & 57.8 &  59.2 & 48.4 \\

		%	\midrule
			
			DeFRCN \cite{defrcn} & ResNet101 & 57.0  &58.6 & 64.3  &67.8 & 67.0 & 35.8  &42.7  &51.0  & 54.4  &52.9  & 52.5  & 56.6  &55.8  &60.7 & 62.5 & 56.0 \\
			
				SNIDA-DeFRCN \cite{snida}& ResNet101 & 59.3 & 60.8 &  64.3 &  65.4  & 65.6  & 35.2  & 40.8  & 50.2  & 54.6  & 50.0  & 51.6  & 52.4  & 55.9  & 58.5  & 62.6  & 55.1 \\
				
				%Norm-VAE &  ResNet101 &  62.1 &  64.9&   67.8 &  69.2 &  67.5&   39.9 &  46.8 &  54.4 &  54.2 &  53.6 &  58.2 &  60.3 &  61.0&   64.0 &  65.5 & 59.3 \\
				
				ECEA \cite{ecea} & ResNet101 & 59.7  & 60.7  & 63.3  & 64.1  & 64.7  & 43.1 &  45.2  & 49.4  & 50.2  & 51.7  & 52.3  & 54.7  & 58.7  & 59.8  & 61.5 & 56.0\\

			\rowcolor{lightgray} 
			 SCSM & ResNet101 &  61.1  &  65.1 &   64.6  &  68.7  &  67.4  &  39.9  &  47.4  &  52.1  &  55.1  &  55.0  &  52.9  &  59.7  &  62.4  &  63.8  &  64.5  &   58.6\\

			%\textbf{60.9} & \textbf{65.1}  & \textbf{64.6}   & \textbf{68.7}  & \textbf{67.4}  			& \textbf{39.9} & \textbf{47.4}  & \textbf{49.9}  & \textbf{55.1}  & \textbf{55.0} 					& \textbf{52.9}  &  \textbf{59.7}   & \textbf{62.4}  & \textbf{63.8}  & \textbf{64.5} & \textbf{58.5} \\
			
			%\midrule
			
			% & MFD & 63.4  & \textbf{66.3}  & 67.7  & 69.4  & 68.1  & 42.1 &  46.5 &  53.4  & \textbf{55.3}  & 53.8  & 56.1 &  58.3  & 59.0  & 62.2  & 63.7 & 58.8 \\
			
			%&	SCSM-MFD & \textbf{63.4} &  65.9  & \textbf{68.1}  &  \textbf{69.7} &  \textbf{69.1} & \textbf{44.6} &  \textbf{51.6}  & \textbf{53.4}  & 54.0  & \textbf{55.1}  & \textbf{56.8}  & \textbf{59.8}  & \textbf{61.4}  & \textbf{63.5}  & \textbf{65.2} & 59.8 \\

			\midrule

	 DeFRCN* \cite{defrcn} & Swin-B  &  66.1   & 69.5   &  70.1   & 74.8   &  74.7   &  53.7    & 54.2   &  55.8   &  60.8   &  61.6   &  54.6   &  61.0   &  64.5   &  68.5   &  68.2 & 63.9 \\

	FM-FSOD \cite{fsod20243} & ViT-B & 40.9  &  52.8  & 59.5  & 68.3  & 71.4  & 33.5  & 36.1  & 48.1  & 53.6  & 59.3  & 41.9 &  52.6 &  54.9  & 62.8  & 68.2 & 53.6\\
	
	FM-FSOD \cite{fsod20243} & ViT-L  & 40.1 &  53.5 &  57.0 &  68.6 &  72.0 &  33.1 &  36.3  & 48.8  & 54.8 &  64.7  & 39.2  & 50.2  & 55.7  &  63.4  &  68.1 & 53.7 \\
	
	DE-ViT \cite{devit} &  ViT-B &  56.9&  61.8 &  68.0 & 73.9 & 72.8 & 45.3 & 47.3 & 58.2 & 59.8 & 60.6 & \textbf{58.6}&  62.3&  62.7&  64.6 & 67.8 & 61.4 \\ 
	
	DE-ViT \cite{devit} & ViT-L  &  55.4  &  56.1 &   68.1 &   70.9  &  71.9  &  43.0  &  39.3  &  58.1 &   61.6 &   63.1  &  58.2  &  \textbf{64.0}  &  61.3  &  64.2  &  67.3  &  60.2
	\\

\rowcolor{lightgray} 
			 SCSM
			  & Swin-B & \textbf{66.8} & \textbf{69.8} & \textbf{73.1} & \textbf{75.5} & \textbf{75.8} 
			 & \textbf{54.0} & \textbf{56.0}  & \textbf{60.1}  & \textbf{62.8}  & \textbf{65.6} &  58.0 & 61.5  & \textbf{65.1}  & \textbf{69.7}  & \textbf{70.8}  & \textbf{65.6} \\

			 \midrule

			% DeFRCN* \cite{defrcn} & Swin-L  & 72.4 & 73.2 & 75.1  & 78.0 & 79.2 & 60.4 & 63.0 & 64.9 & 67.9 & 69.5 & 63.2 & 67.9  & 69.3 & 71.6 &  72.1 \\
			 
			% \rowcolor{lightgray} 
			% SCSM-DeFRCN & Swin-L & \\

		%	\midrule			
			
	\end{tabular}}
	\label{voc}
	%\vspace{-0.3cm}
\end{table*}

	\section{Experiments on the SCSM Module}
	
	In this section, we provide details on the experimental setting of our proposed SCSM module. We then discuss the comparison experiments conducted on SCSM with the latest FSOD methods, as well as ablation studies on SCSM components and parameters.
	
	\subsection{Experimental Setting}
	\textbf{Evaluation Datasets.}
	Following the previous evaluation protocol, we assess the performance of SCSM using two widely recognized FSOD benchmark datasets, namely (VOC07 and VOC12) \cite{voc} and COCO2014 \cite{coco}. According to previous works \cite{TFA,defrcn}, we split the 20 categories into three partitions on the VOC dataset. Each partition contains 15 base classes and 5 novel classes. The evaluation metric utilized is AP50 for 1, 2, 3, 5, and 10 shots. Regarding the COCO dataset, we adopt the 20 categories from VOC as novel classes, with the remaining 60 categories serving as base classes. The results are reported using the AP metric for 10 and 30 samples. It is worth noting that the training and testing of the novel stage are conducted using the data provided by DeFRCN \cite{defrcn}. %All results are reported by averaging over 20 multiple runs. % on the G-FSOD setting.

	\begin{table}[t]
		\centering	
		\tabcolsep=0.1cm
		%\vspace{0.5cm}
		%\renewcommand{\arraystretch}{1.2}
		%\tabcolsep=0.1cm
		%\vspace{-0.5cm}
		%\setlength{\belowcaptionskip}{0.15cm} 
		%\renewcommand{\arraystretch}{1.1}
		
		%	\vspace{-0.3cm}
		\caption{
			Performance comparison among SCSM and mainstream FSOD methods on the COCO dataset.  Symbol `-' represents unreported results in the original work  and  Swin-B is the backbone of Swin Transfromer with large size. % The results are reported by averaging over multiple runs. Bold font indicates the SOTA result in the group.
		}
		\vspace{-0.3cm}
		\scalebox{1}{
			\begin{tabular}{l|c|cc|cc}
				\midrule
				
				\multirow{2}{*}{Methods/shots} &  \multirow{2}{*}{Backbone}& \multicolumn{2}{c|}{10 shots}  & \multicolumn{2}{c}{30 shots}      \\
				
				&   & nAP   & nAP75  & nAP   & nAP75  \\ \midrule

				%	 TFA w/ fc~\cite{TFA}  & ResNet101                                                                                            & 9.1   & 8.5  & 12.1 & 11.8\\
				
				TFA w/ cos~\cite{TFA} & ResNet101  & 10.0 & 8.8 & 13.4 & 12.0\\

			%	& MPSR \cite{MPSR}   & 9.8 & 9.7 & 14.1 & 14.2\\
				
			%	FADI \cite{neurips2021}& ResNet101  & 12.2 & 11.9 & 16.1 & - \\
				
				FSCE~\cite{fsce} & ResNet101                                                                                                  & 11.9 & 10.1&  16.4 & 14.7\\

				FCT~\cite{fct} & PVTv2                                                                                                 & 17.1 & 17.0 & 21.4 & 22.1\\

				VFA  \cite{vfa} & ResNet101 & 15.9 &  - & 18.4 & -\\
				
				%	& Vanilla-VAE  \cite{vae2}  & 18.7  & 22.5 \\
				
				Norm-VAE  \cite{vae2}  & ResNet101 & 18.7 & 17.6  &  22.5 & 22.4\\
				
				%	& Meta-tuning \cite{metatuning}  & 18.8 &\textbf{23.4} \\
				
				%	&ECEA \cite{ecea} &9.6 &13.2 &15.4 &16.7 &19.6 &23.1 \\

				%FPD \cite{fpd}& ResNet101 & 16.5 & -  & 20.1 & -\\

				DeFRCN \cite{defrcn} & ResNet101 &  18.6 &  17.6 & 22.5 & 22.3 \\ 
				
				NIFF \cite{niff}& ResNet101  & 19.1  &  -  & 21.0  & - \\
					
				BSDet \cite{tmm1} & ResNet101  & 17.2  &  -  & 21.2  & - \\

				DAnA \cite{tmm3} & ResNet101  & 18.6  &  17.2  & 21.6  & 20.3 \\

			%	SNIDA-DeFRCN \cite{snida} & ResNet101 &  19.1 & -&  23.1&  -\\
				
				\rowcolor{lightgray}			 
				SCSM & ResNet101 & 19.7 & 18.9 &  23.1   & 23.7 \\
				
				\midrule
				
				DeFRCN* \cite{defrcn} & Swin-B  &  19.4   &  19.2   &  24.3   & 24.5      \\
				
				\rowcolor{lightgray} 
				%SCSM-DeFRCN & Swin-B & \textbf{19.9} & \textbf{19.8} & \textbf{24.7}  & \textbf{24.8}  \\
				
				SCSM & Swin-B & 20.1 & 19.8 & 26.2  & 27.1  \\
				
				\bottomrule
				
				DeFRCN* \cite{defrcn} & Swin-L  &  21.2   &  21.7   &  25.6   & 25.8      \\
				
				\rowcolor{lightgray} 			
				SCSM & Swin-L & \textbf{22.4} & \textbf{23.5} & \textbf{27.8}  & \textbf{28.6}  \\
				
				\bottomrule
				
		\end{tabular}}
		
		\label{COCO}	
		%\vspace{-0.3cm}
	\end{table}

	\textbf{Implementation Details.}
To fairly evaluate the performance of our SCSM, similar to \cite{TFA,ecea}, we utilize Faster R-CNN \cite{fasterrcnn} as our framework and follow fine-tuning strategies from \cite{defrcn}. Additionally, we first employ a pre-trained ResNet101 on ImageNet \cite{imagenet} as our backbone and utilize SGD optimization with momentum and weight decay set to 0.9 and 0.0001, respectively. We then utilize the another backbone that is pre-trained on Swin-Transformer \cite{swin} to assess the stability and adaptability of SCSM. During the base and novel training phases, we conduct training on 4 Nvidia GeForce RTX 3090 GPUs with a learning rate of 0.01 and a batch size of 16. Our novel results are reported by the G-FSOD setting and DeFRCN \cite{defrcn} as our baseline framework.
	%In the base training phase, all modules remain unfrozen. For the novel phase, only the CSS attention layer and RPN module remain unfrozen in the G-FSOD protocol and only the ROI head is frozen on the FSOD protocol. 

	\subsection{Comparison Results}
	
	\textbf{Result on VOC.} For a fair comparison, following \cite{ecea, defrcn}, we evaluate the performance improvement of our SCSM module on three split VOC datasets. We utilize the detection framework provided by DeFRCN with the backbone of ResNet101 as a baseline and plug our SCSM module to report the results. Table \ref{voc} illustrates the results with nAP50 as the metric. From the table, compared with the baseline, our approach demonstrates a significant performance boost. Furthermore, compared to the latest method SNIDA-DeFRCN \cite{snida} with the same backbone of ResNet101, SCSM consistently outperforms them across various splits and shot numbers. Such results serve as a clear indication of the robust capabilities of our module in FSOD tasks, as it effectively enhances the quality of channel feature representation and effectively improves the knowledge transfer and few-shot adaptation.
	
	In addition, we integrate the baseline framework with the backbone of Swin Transformer \cite{swin}, which demonstrates superior performance compared to ResNet101 on the VOC novel classes, as illustrated in Table \ref{voc}(bottom). Although this Transformer-based feature extractor enhances the processing capabilities for few-shot samples, it still exhibits limitations in data-scarce scenarios, failing to highlight the effective patterns and rectifying ineffective ones in feature channels. As for this challenge, our SCSM not only effectively integrates with the Swin Transformer but also significantly boosts the overall performance of FSOD by enhancing the quality of channel features.
	
	\textbf{Result on COCO.} Our SCSM follows training strategies and parameters that align with VOC on the COCO dataset. The comparative results on COCO are presented in Table \ref{COCO}. It is observed that our method has achieved a 1.1\% improvement of nAP in the 10-shot setting compared to the baseline DeFRCN \cite{defrcn}. Additionally, our method demonstrates significant advantages over the latest approaches. Such results convincingly demonstrate that SCSM effectively enhances FSOD performance through channel feature modeling. On the other hand, our method demonstrates consistent effectiveness on the COCO dataset when applied to the Swin Transformer backbone. The experimental results from both VOC and COCO datasets indicate that SCSM not only enhances the  quality of channel features but also exhibits strong generalization capabilities across different datasets and backbones.
		\begin{figure*}[t]
		\begin{center}
			%\fontsize{8}{11}\selectfont 
			\includegraphics[width=1\textwidth]{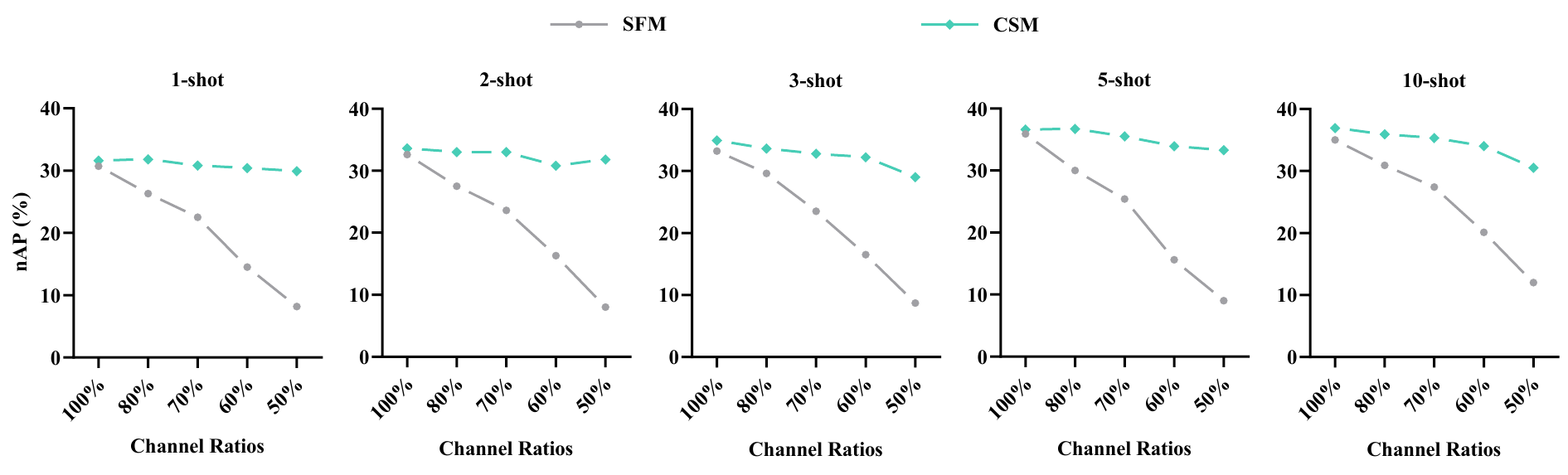}
		\end{center}
		\setlength{\abovecaptionskip}{-0.1cm} 
		\caption{Enhancing the quality of channel features via spatial-channel state space modeling.
		}
		%	\vspace{-0.3cm} 
		\label{channelratio}
		%\vspace{-0.2cm} 
	\end{figure*}

	\textbf{Effectiveness of SCSM Module on the Traditional Object Detection.}
Given that the base model is trained on a large annotated dataset, we employ SCSM to enhance the feature quality of the base class and examine its impact on the performance of traditional object detection. To verify this, Table \ref{bap} shows the performance enhancements achieved by SCSM in generic object detection through the learning of the base class on the VOC and COCO datasets. The table demonstrates that, when compared to two baseline models \cite{TFA,defrcn}, SCSM offers a modest performance improvement over FSOD. For instance, using our baseline model \cite{defrcn}, we observe enhancements from 81.0 to 81.2 of the average results on three VOC-base datasets (as shown in Table \ref{bap}) and from 56.0 to 58.6 of average results on the VOC dataset (refer to Table \ref{voc}). The abundance of labeled datasets containing rich category features aids the model in learning a more accurate representation of data distributions, allowing for more precise and effective channel feature representation. However, the high quality of these feature representations somewhat restricts the potential improvements of SCSM. \textbf{Therefore, the results from the base classes suggest that the SCSM module is more advantageous for the FSOD task compared to traditional object detection.}

			\begin{table}[t]
		\centering	
		\tabcolsep=0.2cm
		\caption{
			Impact of SCSM on traditional object detection performance. We use bAP50 as the evaluation metric.
		}
		\vspace{-0.3cm}
		\scalebox{1}{			
			\begin{tabular}{l|cccc|c}
				\midrule
				
				\multicolumn{1}{c|}{\multirow{2}{*}{Methods}} & \multicolumn{4}{c|}{VOC} &\multicolumn{1}{c}{\multirow{2}{*}{COCO}}       \\

				& Set1   & Set2 & Set3 & Avg.  \\ \midrule
				
				TFA \cite{TFA} & 80.8  & 81.2 & 81.4 & 81.1 & -\\
				
				DeFRCN \cite{defrcn}  & 80.3  & 81.7  & 81.1 & 81.0 & 59.2 \\
				
				%	& ECEA  \cite{vfa}  & 82.1  & 82.4  & 82.8 & 60.2\\
				
				SCSM (Ours) & 80.8  & 81.7 & 81.1 & 81.2 & 59.8\\

				\bottomrule
				
		\end{tabular}}
		
		\label{bap}	
		%\vspace{-0.3cm}
	\end{table}

	% \begin{figure*}[t]
		%     \begin{center}
			%         %\fontsize{8}{11}\selectfont 
			%         \includegraphics[width=1\textwidth]{figure/f3.pdf}
			%     \end{center}
		%     \setlength{\abovecaptionskip}{-0.1cm} 
		%     \caption{The destruction of Gaussian noise on channel state relationships during testing. We adopt DeFRCN \cite{defrcn} as our baseline and report nAP results on VOC-split1 at the 1-shot setting.}
		%     %\vspace{-0.3cm} 
		%     \label{f3}
		%     %\vspace{-0.2cm} 
		% \end{figure*}

	\subsection{Ablation Study}

\textbf{Can CSM Improve the Quality of Channel Features?}
To validate this phenomenon, we conduct a comprehensive experimental analysis on the VOC-Split1 dataset to systematically evaluate the model's performance under varying proportions of high-weight channels. Specifically, we assess the model by selectively extracting channels with weights in the top 80\%, 70\%, 60\%, and 50\%, where the channel weights are derived using SENet. The experimental results, as illustrated in Fig. \ref{channelratio}, present the performance metrics in terms of nAP. The line charts reveal a significant trend: as the number of channel features decreases, the CSM model exhibits minimal performance degradation across all shot settings, demonstrating remarkable robustness. In contrast, SFM, which lacks channel state modeling, experiences a substantial decline in performance as the number of significant channel features is reduced. This outcome highlights the ability of the SCSM model to maintain the quality of a larger proportion of channel features, thereby effectively mitigating performance degradation in the FSOD task. Consequently, this experimental result demonstrates that SCSM enables the model to accurately emphasize effective channels while rectifying those incorrect ones that the model might otherwise focus on.

	\textbf{Performance of SFM and CSM Components.}
	To validate the respective improvements of SFM and CSM on FSOD performance, we conduct an ablation study on the VOC-split1 dataset. Table \ref{ab} illustrates the results. From the table, CSM exceeds the baseline in all shot settings, especially in the 1-shot and 2-shot scenarios, with 2.1\% and 5.5\% achieved over the baseline, respectively. This indicates that CSM has a significant effect in capturing channel features and effectively enhancing the model's feature representation capability. In addition, inspired by spatial-channel attention \cite{cbam,eca}, we introduce SFM to balance the feature learning between spatial and channel aspects.
	Using SFM alnoe outperforms the baseline under each shot setting too. Notably, the comparative analysis reveals that CSM outperforms SFM, highlighting that the novel model demonstrates relatively weaker capabilities in channel feature modeling compared to spatial feature modeling. This observation emphasizes the crucial role of channel modeling in FSOD tasks. More importantly, when introducing both SFM and CSM components, the model's performance reaches an optimal level. Such results suggest that balanced spatial-channel feature modeling can significantly improve FSOD performance.

	\begin{table}[t]
		\centering
		\tabcolsep=0.25cm
		\caption{
			Performance of SFM and CSM components. }
		\vspace{-0.2cm}
		\scalebox{1}{
			\begin{tabular}{cc|ccccc}
				\hline			
				
				\multicolumn{1}{c}{\multirow{2}{*}{CSM}} & \multicolumn{1}{c|}{\multirow{2}{*}{SFM}} & \multicolumn{5}{c}{Shot Number}                                                                                           \\  
				
				& & 1    &2 &5    & 10       & Avg.           \\  \hline
				
				& & 57.0 & 58.6 & 67.8 & 67.0 & 62.6 \\
				%\CheckmarkBold &  & 58.5  & 64.5  & 68.1  & 67.0  &  64.5\\
			\CheckmarkBold	&  & 59.1 & 64.1  &68.5 & 67.9  & 64.9 \\
				\CheckmarkBold &\CheckmarkBold  &  \textbf{61.1} & \textbf{65.1}  & \textbf{68.7}   & \textbf{67.4}  &  \textbf{65.6}                   \\

				\hline
		\end{tabular}}

		\label{ab}
		%\vspace{-0.3cm}
	\end{table}

	\textbf{Performance of Different Spatial or Channel Modeling Module in FSOD Tasks.}
	We introduce a CNN-based Adapter \cite{cnnadapter} after the backbone to indirectly facilitate spatial feature modeling while directly increasing the network's depth, with the aim of enhancing the model's performance. However, as indicated in Table \ref{pa}, this increase in depth does not lead to performance improvements compared to the baseline. It may even heighten the risk of overfitting. We then directly integrate VIM \cite{vim}, as the mamba-based vision backbone, into the DeFRCN framework for spatial feature modeling to evaluate the performance of the mamba-based feature extractor on the FSOD task. From Table \ref{pa}, VIM demonstrates an average performance that surpasses the baseline. This indicates that VIM enhances FSOD performance through spatial feature modeling.	
	Furthermore, channel attention-based methods can improve the performance of FSOD by effectively channel feature modeling. As demonstrated in Table \ref{pa}, incorporating channel attention techniques, e.g., SENet \cite{senet}, CBAM \cite{cbam}, ECA \cite{eca}, Self-Attention \cite{self-attention}, and our proposed CSM into the baseline model leads to enhanced FSOD performance, especially in the 1-shot and 2-shot settings. This experimental result indicates that as data becomes scarcer, the number of invalid and redundant feature channels increases. Noteworthy, the final results indicate that, compared with existing channel attention methods \cite{senet,cbam,eca}, CSM is the most effective channel feature modeling technique for FSOD tasks.

	\begin{table}[t]
		\centering
		\tabcolsep=0.2cm
		\caption{
			Performance of different spatial or channel modeling modules on the VOC-Split1 dataset. Except for the baseline \cite{defrcn}, all results are reported by us on the same seed. }
		\vspace{-0.3cm}
		\scalebox{1}{
			\begin{tabular}{l|ccccc}
					\hline
					
				\multicolumn{1}{c|}{\multirow{2}{*}{Methods}}  &		 \multicolumn{5}{c}{Shot Number}                                                                                           \\ 
				& 1             & 2             & 5             & 10                      & \multicolumn{1}{c}{Avg.} \\ \hline 
				
				DeFRCN \cite{defrcn} & 57.0 & 58.6 & 67.8 & 67.0 & 62.6 \\ 
				
				\hline

	 			CNNAdapter \cite{cnnadapter}                        & 53.8          & 57.5          & 64.0          & 63.6       &  59.7                       \\ 
				
				VIM \cite{vim}  & 58.9 & 61.5 & 67.9 &  66.1 & 63.6  \\

				%ECEA \cite{ecea} & 59.7  & 60.7    & 64.1  & 64.7 & \\

					SENet \cite{senet} & 58.3  & 63.9 &  66.4  & 66.1 & 63.7 \\
				 
					CBAM \cite{cbam} & 56.7  & 62.3  & 66.7  & 65.7 & 62.9 \\
				
					ECA-Net \cite{eca}  &  56.7 &  63.4 &  66.9 &  65.1  & 63.0
				\\ 
			%	\rowcolor{lightgray}
			
		Self-Attention \cite{self-attention}	& 58.5  & 63.5  & 67.2  & 67.1  & 64.1  \\

			CSM (Ours) & \textbf{59.1} & \textbf{64.1}  &\textbf{68.5} & \textbf{67.9}  & \textbf{64.9} \\

%	\multirow{4}{*}{(b)}	& 	SFM   & 58.5  & 64.5  & 68.1  & 67.0  &  64.5 \\ 

			%	& 	SFM+SENet  \cite{senet}  & 60.5          & 64.7          & 67.9          & 66.0    &   64.7     \\
				
		%		& 	SFM+SA  \cite{self-attention} & 59.8    & \textbf{65.5}         & 67.5          & 67.1            & 65.0  \\  
		%		\rowcolor{lightgray}
		%		& 	SFM+CSM  &  \textbf{61.1} & 65.1  & \textbf{68.7}   & \textbf{67.4}  &  \textbf{65.6}       \\  	

				\hline 
		\end{tabular}}
		%\vspace{1mm}
		
		\label{pa}
			\vspace{-0.3cm}
	\end{table}

%	We utilize Grad-CAM to visualize the novel objects on the VOC-split1 (10 shots) datasets, as illustrated in Figure \ref{heatmap}. From the figure, the heatmap visualization reveals significant confusion between background and foreground attention in DeFRCN \cite{defrcn}, particularly in complex scenes. In contrast, our SCSM accurately captures the key features of the object, alleviating the confusion between background and foreground.  This indicates that our approach effectively enhances feature representation through channel state relationship modeling, improving the knowledge transfer and few-shot adaptation.
	
%	Furthermore, the baseline model experiences issues with catastrophic forgetting, leading to confusion regarding certain features of the base classes during the fine-tuning stage. Fortunately, our SCSM effectively addresses this limitation.
\section{Visualization Analysis}

\subsection{Visualization on the Feature Level}
\begin{figure}[t]
	\vspace{0.3cm}
	\begin{center}
		%\fontsize{8}{11}\selectfont 
		\includegraphics[width=0.5\textwidth]{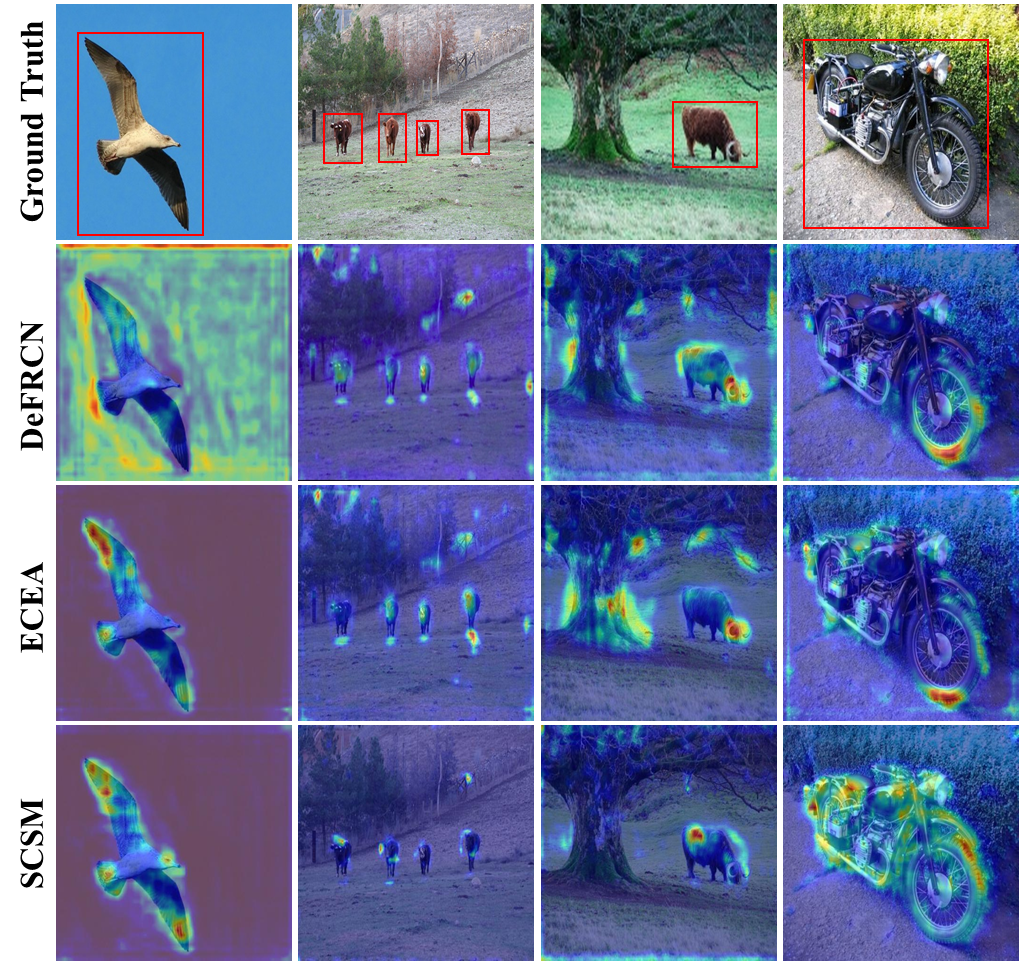}
	\end{center}
	\setlength{\abovecaptionskip}{-0.1cm} 
	\caption{Heatmap visualization of novel objects on the VOC-split1 test dataset. 
	}

	\label{heatmap}
 
\end{figure}

We utilize Grad-CAM to visualize novel objects in the VOC-split1 (10 shots) dataset, as illustrated in Fig. \ref{heatmap}. The resulting heatmap highlights considerable attention confusion between the background and foreground in both the DeFRCN model \cite{defrcn} and ECEA \cite{ecea}, particularly within complex scenes. The redundancy and insufficient feature information in the channels ultimately impede the FSOD model's ability to learn feature correlations effectively. In contrast, our SCSM adeptly captures the essential features of the objects, alleviating the confusion between background and foreground. This indicates that our approach significantly enhances feature representation through effective channel state relationship modeling.

\subsection{Detection Visualization}
\begin{figure}[t]
	\begin{center}
		%\fontsize{8}{11}\selectfont 
		\includegraphics[width=0.5\textwidth]{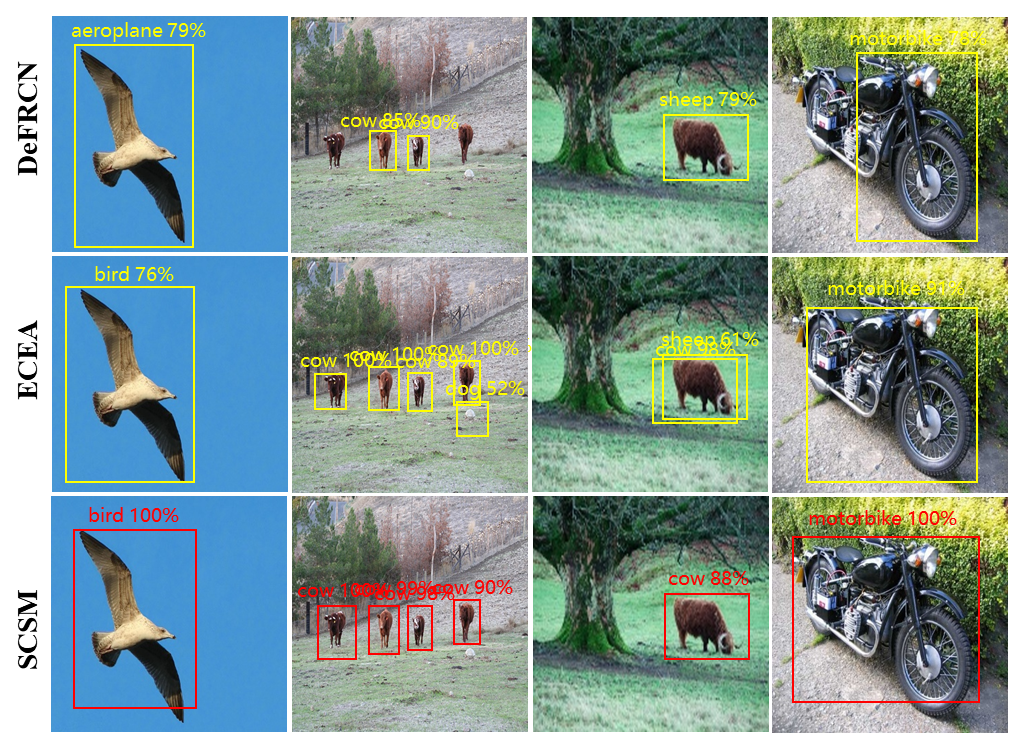}
	\end{center}
	
	\setlength{\abovecaptionskip}{-0.1cm} 
	\caption{Detection visualization on the VOC-Split1 test dataset.		
	}
	\vspace{-0.3cm} 
	\label{visbox}
	%\vspace{-0.2cm} 
\end{figure}

We present the detection results for the above images in Fig. \ref{visbox}. The figure reveals that DeFRCN \cite{defrcn} struggles with detection omission and classification confusion, primarily due to its reliance on redundant and erroneous class-specific features. Although ECEA employs spatial feature modeling to improve detection performance, instances of repeated and incorrect detections still occur within the images. In contrast, our SCSM effectively captures the essential fine-grained features of objects, thereby reducing confusion between background and foreground. This demonstrates that our method significantly enhances feature representation by modeling channel state relationships, ultimately improving the performance of FSOD.

	\section{Conclusion}
	
	In this paper, we considered that, due to the limited availability of data samples, existing FSOD models tended to extract ineffective or redundant channel features when dealing with novel classes. To solve this problem, we proposed an SCSM module, as a variant of Mamba, to handle the semantic gap between base and novel classes by highlighting the correctly transferred patterns and rectifying those incorrect ones in feature channels. Specifically, in SCSM, we designed SFM to ensure that the subsequent extracted channel features are valid and then introduced CSM based on Mamba to learn feature state correlation in channels. Extensive experiments on the VOC and COCO datasets have shown that SCSM enables the novel detector to improve the quality of focused feature representation in channels and enhance the performance of FSOD.
	
	%In this paper, we have proposed an SCSM module, as a variant of Mamba, to handle the semantic gap between base and novel classes by highlighting the correctly transferred patterns and rectifying those incorrect ones in feature channels. Specifically, in SCSM, we designed SFM to ensure that the subsequent extracted channel features are valid and then introduced CSM based on Mamba to learn feature state correlation in channels. Extensive experiments on the VOC and COCO datasets have shown that SCSM enables the novel detector to improve the quality of focused feature representation in channels and enhance the performance of FSOD.
	
%	{
	%	\small
	%	\bibliographystyle{ieeenat_fullname}
	%	\bibliography{SCSM}
	%}
	
	\bibliographystyle{IEEEtran}
	\bibliography{SCSM}

\begin{thebibliography}{10}
\providecommand{\url}[1]{#1}
\csname url@samestyle\endcsname
\providecommand{\newblock}{\relax}
\providecommand{\bibinfo}[2]{#2}
\providecommand{\BIBentrySTDinterwordspacing}{\spaceskip=0pt\relax}
\providecommand{\BIBentryALTinterwordstretchfactor}{4}
\providecommand{\BIBentryALTinterwordspacing}{\spaceskip=\fontdimen2\font plus
\BIBentryALTinterwordstretchfactor\fontdimen3\font minus
  \fontdimen4\font\relax}
\providecommand{\BIBforeignlanguage}[2]{{%
\expandafter\ifx\csname l@#1\endcsname\relax
\typeout{** WARNING: IEEEtran.bst: No hyphenation pattern has been}%
\typeout{** loaded for the language `#1'. Using the pattern for}%
\typeout{** the default language instead.}%
\else
\language=\csname l@#1\endcsname
\fi
#2}}
\providecommand{\BIBdecl}{\relax}
\BIBdecl

\bibitem{tmm1}
Y.~Lu, X.~Chen, Z.~Wu, M.~Tan, and J.~Yu, ``Binary similarity few-shot object
  detection with modeling of hard negative samples,'' \emph{IEEE Transactions
  on Multimedia}, vol.~26, pp. 4805--4818, 2023.

\bibitem{tmm2}
X.~Zhao, X.~Liu, Y.~Ma, S.~Bai, Y.~Shen, Z.~Hao, and A.~Liu, ``Temporal
  speciation network for few-shot object detection,'' \emph{IEEE Transactions
  on Multimedia}, vol.~25, pp. 8267--8278, 2023.

\bibitem{tmm3}
T.-I. Chen, Y.-C. Liu, H.-T. Su, Y.-C. Chang, Y.-H. Lin, J.-F. Yeh, W.-C. Chen,
  and W.~H. Hsu, ``Dual-awareness attention for few-shot object detection,''
  \emph{IEEE Transactions on Multimedia}, vol.~25, pp. 291--301, 2021.

\bibitem{smile}
A.~Majee, R.~Sharp, and R.~Iyer, ``Smile: Leveraging submodular mutual
  information for robust few-shot object detection,'' \emph{arXiv preprint
  arXiv:2407.02665}, 2024.

\bibitem{defrcn}
L.~Qiao, Y.~Zhao, Z.~Li, X.~Qiu, J.~Wu, and C.~Zhang,
  ``\BIBforeignlanguage{English}{Defrcn: Decoupled faster r-cnn for few-shot
  object detection},'' in \emph{\BIBforeignlanguage{English}{Proc. ICCV}},
  Virtual, Online, Canada, 2021, pp. 8661--8670.

\bibitem{fsod20242}
K.~Guirguis, G.~Eskandar, M.~Wang, M.~Kayser, E.~Monari, B.~Yang, and
  J.~Beyerer, ``Uncertainty-based forgetting mitigation for generalized
  few-shot object detection,'' in \emph{Proc. CVPR}, 2024, pp. 2586--2595.

\bibitem{ecea}
Z.~Xin, T.~Wu, S.~Chen, Y.~Zou, L.~Shao, and X.~You, ``Ecea: Extensible
  co-existing attention for few-shot object detection,'' \emph{IEEE
  Transactions on Image Processing}, 2024.

\bibitem{fsod20243}
G.~Han and S.-N. Lim, ``Few-shot object detection with foundation models,'' in
  \emph{Proc. CVPR}, 2024, pp. 28\,608--28\,618.

\bibitem{aft}
T.~Wu, Z.~Xin, S.~Chen, Y.~Zou, and X.~You, ``Adversarial feature training for
  few-shot object detection,'' \emph{IEEE Transactions on Circuits and Systems
  for Video Technology}, pp. 1--1, 2025.

\bibitem{snida}
Y.~Wang, X.~Zou, L.~Yan, S.~Zhong, and J.~Zhou, ``Snida: Unlocking few-shot
  object detection with non-linear semantic decoupling augmentation,'' in
  \emph{Proc. CVPR}, June 2024, pp. 12\,544--12\,553.

\bibitem{senet}
J.~Hu, L.~Shen, and G.~Sun, ``Squeeze-and-excitation networks,'' in \emph{Proc.
  CVPR}, 2018, pp. 7132--7141.

\bibitem{mamba}
A.~Gu and T.~Dao, ``Mamba: Linear-time sequence modeling with selective state
  spaces,'' \emph{arXiv preprint arXiv:2312.00752}, 2023.

\bibitem{cbam}
S.~Woo, J.~Park, J.-Y. Lee, and I.~S. Kweon, ``Cbam: Convolutional block
  attention module,'' in \emph{Proc. ECCV}, 2018, pp. 3--19.

\bibitem{eca}
Q.~Wang, B.~Wu, P.~Zhu, P.~Li, W.~Zuo, and Q.~Hu, ``Eca-net: Efficient channel
  attention for deep convolutional neural networks,'' in \emph{Proc. CVPR},
  2020, pp. 11\,534--11\,542.

\bibitem{self-attention}
A.~Vaswani, N.~Shazeer, N.~Parmar, J.~Uszkoreit, L.~Jones, A.~N. Gomez,
  {\L}.~Kaiser, and I.~Polosukhin, ``Attention is all you need,'' \emph{Proc.
  NeurIPS}, vol.~30, 2017.

\bibitem{xinservey}
Z.~Xin, S.~Chen, T.~Wu, Y.~Shao, W.~Ding, and X.~You, ``Few-shot object
  detection: Research advances and challenges,'' \emph{Information Fusion}, p.
  102307, 2024.

\bibitem{DC}
B.-B. Gao, X.~Chen, Z.~Huang, C.~Nie, J.~Liu, J.~Lai, G.~JIANG, X.~Wang, and
  C.~Wang, ``Decoupling classifier for boosting few-shot object detection and
  instance segmentation,'' in \emph{Proc. NeurIPS}, S.~Koyejo, S.~Mohamed,
  A.~Agarwal, D.~Belgrave, K.~Cho, and A.~Oh, Eds.\hskip 1em plus 0.5em minus
  0.4em\relax Curran Associates, Inc., 2022, pp. 18\,640--18\,652.

\bibitem{EME}
C.~Liu, B.~Li, M.~Shi, X.~Chen, Q.~Ye, and X.~Ji, ``Explicit margin equilibrium
  for few-shot object detection,'' \emph{IEEE Transactions on Neural Networks
  and Learning Systems}, pp. 1--13, 2024.

\bibitem{PCD}
B.~Li, C.~Liu, M.~Shi, X.~Chen, X.~Ji, and Q.~Ye, ``Proposal distribution
  calibration for few-shot object detection,'' \emph{IEEE Transactions on
  Neural Networks and Learning Systems}, vol.~36, no.~1, pp. 1911--1918, 2025.

\bibitem{fsod20241}
Q.~Fan, W.~Zhuo, C.-K. Tang, and Y.-W. Tai, ``Fsodv2: A deep calibrated
  few-shot object detection network,'' \emph{International Journal of Computer
  Vision}, pp. 1--20, 2024.

\bibitem{metafrcn}
G.~Han, S.~Huang, J.~Ma, Y.~He, and S.-F. Chang,
  ``\BIBforeignlanguage{English}{Meta faster r-cnn: Towards accurate few-shot
  object detection with attentive feature alignment. arxiv},'' 2021, arXiv
  preprint arXiv:2104.07719.

\bibitem{metarcnn}
X.~Yan, Z.~Chen, A.~Xu, X.~Wang, X.~Liang, and L.~Lin,
  ``\BIBforeignlanguage{English}{Meta r-cnn: Towards general solver for
  instance-level low-shot learning},'' in
  \emph{\BIBforeignlanguage{English}{Proc. ICCV}}, Los Alamitos, CA, USA, 2019,
  pp. 9576--9585.

\bibitem{qa}
G.~Han, Y.~He, S.~Huang, J.~Ma, and S.-F. Chang,
  ``\BIBforeignlanguage{English}{Query adaptive few-shot object detection with
  heterogeneous graph convolutional networks},'' in
  \emph{\BIBforeignlanguage{English}{Proc. ICCV}}, Virtual, Online, Canada,
  2021, pp. 3243--3252.

\bibitem{TFA}
X.~Wang, T.~E. Huang, T.~Darrell, J.~E. Gonzalez, and F.~Yu,
  ``\BIBforeignlanguage{English}{Frustratingly simple few-shot object
  detection},'' in \emph{\BIBforeignlanguage{English}{Proc. ICML}}, Virtual,
  Online, 2020, pp. 9861--9870.

\bibitem{SRR}
C.~Zhu, F.~Chen, U.~Ahmed, Z.~Shen, and M.~Savvides,
  ``\BIBforeignlanguage{English}{Semantic relation reasoning for shot-stable
  few-shot object detection},'' in \emph{\BIBforeignlanguage{English}{Proc.
  CVPR}}, Piscataway, NJ, USA, 2021, pp. 8778--8787.

\bibitem{niff}
K.~Guirguis, J.~Meier, G.~Eskandar, M.~Kayser, B.~Yang, and J.~Beyerer, ``Niff:
  Alleviating forgetting in generalized few-shot object detection via neural
  instance feature forging,'' in \emph{Proc. CVPR}, June 2023, pp.
  24\,193--24\,202.

\bibitem{fsce}
B.~Sun, B.~Li, S.~Cai, Y.~Yuan, and C.~Zhang,
  ``\BIBforeignlanguage{English}{Fsce: Few-shot object detection via
  contrastive proposal encoding},'' in \emph{\BIBforeignlanguage{English}{Proc.
  CVPR}}, Piscataway, NJ, USA, 2021, pp. 7348--7358.

\bibitem{zhang}
L.~Zhang, S.~Zhou, J.~Guan, and J.~Zhang, ``Accurate few-shot object detection
  with support-query mutual guidance and hybrid loss,'' in \emph{Proc. CVPR},
  2021, pp. 14\,419--14\,427.

\bibitem{repmet}
L.~Karlinsky, J.~Shtok, S.~Harary, E.~Schwartz, A.~Aides, R.~Feris, R.~Giryes,
  and A.~Bronstein, ``\BIBforeignlanguage{English}{Repmet: representative-based
  metric learning for classification and few-shot object detection},'' in
  \emph{\BIBforeignlanguage{English}{Proc. CVPR}}, Los Alamitos, CA, USA, 2019,
  pp. 5192--5201.

\bibitem{metricfsod2}
Y.~Li, W.~Feng, S.~Lyu, and Q.~Zhao, ``\BIBforeignlanguage{English}{Feature
  reconstruction and metric based network for few-shot object detection},''
  \emph{\BIBforeignlanguage{English}{Computer Vision and Image Understanding}},
  pp. 103\,600--103\,610, 2023.

\bibitem{vfa}
J.~Han, Y.~Ren, J.~Ding, K.~Yan, and G.-S. Xia,
  ``\BIBforeignlanguage{English}{Few-shot object detection via variational
  feature aggregation. arxiv},'' 2023, arXiv preprint at arXiv.2301.13411.

\bibitem{vae2}
J.~Xu, H.~Le, and D.~Samaras, ``Generating features with increased crop-related
  diversity for few-shot object detection,'' in \emph{Proc. CVPR}, 2023, pp.
  19\,713--19\,722.

\bibitem{fct}
G.~Han, J.~Ma, S.~Huang, L.~Chen, and S.-F. Chang,
  ``\BIBforeignlanguage{English}{Few-shot object detection with fully
  cross-transformer},'' in \emph{\BIBforeignlanguage{English}{Proc. CVPR}},
  Piscataway, NJ, USA, 2022, pp. 5311--5320.

\bibitem{metadetr}
G.~Zhang, Z.~Luo, K.~Cui, S.~Lu, and E.~P. Xing,
  ``\BIBforeignlanguage{English}{Meta-detr: Image-level few-shot detection with
  inter-class correlation exploitation},''
  \emph{\BIBforeignlanguage{English}{IEEE Transactions on Pattern Analysis and
  Machine Intelligence}}, pp. 1--12, 2022.

\bibitem{mamba2}
T.~Dao and A.~Gu, ``Transformers are ssms: Generalized models and efficient
  algorithms through structured state space duality,'' \emph{arXiv preprint
  arXiv:2405.21060}, 2024.

\bibitem{videomamba}
K.~Li, X.~Li, Y.~Wang, Y.~He, Y.~Wang, L.~Wang, and Y.~Qiao, ``Videomamba:
  State space model for efficient video understanding,'' \emph{arXiv preprint
  arXiv:2403.06977}, 2024.

\bibitem{mambasurvey}
R.~Xu, S.~Yang, Y.~Wang, B.~Du, and H.~Chen, ``A survey on vision mamba:
  Models, applications and challenges,'' \emph{arXiv preprint
  arXiv:2404.18861}, 2024.

\bibitem{vim}
L.~Zhu, B.~Liao, Q.~Zhang, X.~Wang, W.~Liu, and X.~Wang, ``Vision mamba:
  Efficient visual representation learning with bidirectional state space
  model,'' \emph{arXiv preprint arXiv:2401.09417}, 2024.

\bibitem{vmamba}
Y.~Liu, Y.~Tian, Y.~Zhao, H.~Yu, L.~Xie, Y.~Wang, Q.~Ye, and Y.~Liu, ``Vmamba:
  Visual state space model,'' \emph{arXiv preprint arXiv:2401.10166}, 2024.

\bibitem{localmamba}
T.~Huang, X.~Pei, S.~You, F.~Wang, C.~Qian, and C.~Xu, ``Localmamba: Visual
  state space model with windowed selective scan,'' \emph{arXiv preprint
  arXiv:2403.09338}, 2024.

\bibitem{fcanet}
Z.~Qin, P.~Zhang, F.~Wu, and X.~Li, ``Fcanet: Frequency channel attention
  networks,'' in \emph{Proc. ICCV}, 2021, pp. 783--792.

\bibitem{gcnet}
Y.~Cao, J.~Xu, S.~Lin, F.~Wei, and H.~Hu, ``Global context networks,''
  \emph{IEEE Transactions on Pattern Analysis and Machine Intelligence},
  vol.~45, no.~6, pp. 6881--6895, 2020.

\bibitem{resnet}
K.~He, X.~Zhang, S.~Ren, and J.~Sun, ``Deep residual learning for image
  recognition,'' in \emph{Proc. CVPR}, 2016, pp. 770--778.

\bibitem{swin}
Z.~Liu, Y.~Lin, Y.~Cao, H.~Hu, Y.~Wei, Z.~Zhang, S.~Lin, and B.~Guo,
  ``\BIBforeignlanguage{English}{Swin transformer: Hierarchical vision
  transformer using shifted windows},'' in
  \emph{\BIBforeignlanguage{English}{Proc. ICCV}}, Virtual, Online, Canada,
  2021, pp. 9992--10\,002.

\bibitem{hippo}
A.~Gu, I.~Johnson, A.~Timalsina, A.~Rudra, and C.~R{\'e}, ``How to train your
  hippo: State space models with generalized orthogonal basis projections,''
  \emph{arXiv preprint arXiv:2206.12037}, 2022.

\bibitem{metadet}
Y.-X. Wang, D.~Ramanan, and M.~Hebert,
  ``\BIBforeignlanguage{English}{Meta-learning to detect rare objects},'' in
  \emph{\BIBforeignlanguage{English}{Proc. ICCV}}, Los Alamitos, CA, USA, 2019,
  pp. 9924--9933.

\bibitem{fpd}
Z.~Wang, B.~Yang, H.~Yue, and Z.~Ma, ``Fine-grained prototypes distillation for
  few-shot object detection,'' in \emph{Proc. AAAI}, vol.~38, no.~6, 2024, pp.
  5859--5866.

\bibitem{devit}
X.~Zhang, Y.~Liu, Y.~Wang, and A.~Boularias, ``Detect everything with few
  examples,'' \emph{arXiv preprint arXiv:2309.12969}, 2023.

\bibitem{voc}
M.~Everingham, L.~Van~Gool, C.~Williams, J.~Winn, and A.~Zisserman,
  ``\BIBforeignlanguage{English}{The pascal visual object classes (voc)
  challenge},'' \emph{\BIBforeignlanguage{English}{International Journal of
  Computer Vision}}, pp. 303--308, 2010.

\bibitem{coco}
T.-Y. Lin, M.~Maire, S.~Belongie, J.~Hays, P.~Perona, D.~Ramanan, P.~Dollar,
  and C.~Zitnick, ``\BIBforeignlanguage{English}{Microsoft coco: Common objects
  in context},'' in \emph{\BIBforeignlanguage{English}{Proc. ECCV}}, Cham,
  Switzerland, 2014, pp. 740--755.

\bibitem{fasterrcnn}
S.~Ren, K.~He, R.~Girshick, and J.~Sun, ``Faster r-cnn: Towards real-time
  object detection with region proposal networks,'' \emph{IEEE Transactions on
  Pattern Analysis and Machine Intelligence}, vol.~39, no.~6, pp. 1137--1149,
  2017.

\bibitem{imagenet}
O.~Russakovsky, J.~Deng, H.~Su, J.~Krause, S.~Satheesh, S.~Ma, Z.~Huang,
  A.~Karpathy, A.~Khosla, M.~Bernstein, A.~Berg, and L.~Fei-Fei,
  ``\BIBforeignlanguage{English}{Imagenet large scale visual recognition
  challenge},'' \emph{\BIBforeignlanguage{English}{International Journal of
  Computer Vision}}, pp. 211--252, 2015.

\bibitem{cnnadapter}
J.~Pfeiffer, A.~Kamath, A.~R{\"u}ckl{\'e}, K.~Cho, and I.~Gurevych,
  ``Adapterfusion: Non-destructive task composition for transfer learning,''
  \emph{arXiv preprint arXiv:2005.00247}, 2020.

\end{thebibliography}
	

\end{document}